\documentclass[conference, 10pt]{IEEEtran}
\usepackage{cite}
\usepackage{amsmath,amssymb,amsfonts}
\usepackage{mathtools}
\usepackage{algorithmic}
\usepackage{textcomp}
\usepackage{booktabs}
\usepackage{multirow}
\usepackage{graphicx}
\usepackage[table,xcdraw]{xcolor}
\usepackage{hyperref}
\usepackage{svg}
\usepackage[caption=false]{subfig}
\usepackage[a4paper, left=13mm, right=13mm, top=19mm, bottom=43mm]{geometry}
\def\BibTeX{{\rm B\kern-.05em{\sc i\kern-.025em b}\kern-.08em
    T\kern-.1667em\lower.7ex\hbox{E}\kern-.125emX}}
\DeclarePairedDelimiter{\ceil}{\lceil}{\rceil}
\begin{document}

\title{Effective Analog ICs Floorplanning with Relational Graph Neural Networks and Reinforcement Learning}


\author{
\IEEEauthorblockN{1\textsuperscript{st}
Davide Basso
2\textsuperscript{nd}
Luca Bortolussi
}
\IEEEauthorblockA{\textit{University of Trieste}\\
Trieste, Italy\\
davide.basso@phd.units.it,
lbortolussi@units.it
}
\and
\IEEEauthorblockN{3\textsuperscript{rd}
Mirjana Videnovic-Misic
\IEEEauthorblockA{\textit{Infineon Technologies AT}\\
Villach, Austria\\
mirjana.videnovic-misic@infineon.com}}
\and
\IEEEauthorblockN{4\textsuperscript{th}
Husni Habal
\IEEEauthorblockA{\textit{Infineon Technologies AG}\\
Munich, Germany\\
husni.habal@infineon.com}}
}

\makeatletter
\newcommand{\linebreakand}{%
  \end{@IEEEauthorhalign}
  \hfill\mbox{}\par
  \mbox{}\hfill\begin{@IEEEauthorhalign}
}
\makeatother
\bstctlcite{IEEEexample:BSTcontrol}

\maketitle

\begin{abstract}
Analog integrated circuit (IC) floorplanning is typically a manual process with the placement of components (devices and modules) planned by a layout engineer. This process is further complicated by the interdependence of floorplanning and routing steps, numerous electric and layout-dependent constraints, as well as the high level of customization expected in analog design.
This paper presents a novel automatic floorplanning algorithm based on reinforcement learning. It is augmented by a relational graph convolutional neural network model for encoding circuit features and positional constraints.
The combination of these two machine learning methods enables knowledge transfer across different circuit designs with distinct topologies and constraints, increasing the \emph{generalization ability} of the solution. 
Applied to $6$ industrial circuits, our approach surpassed established floorplanning techniques in terms of speed, area and half-perimeter wire length. When integrated into a \emph{procedural generator} for layout completion, overall layout time was reduced by $67.3\%$ with a $8.3\%$ mean area reduction compared to manual layout.
\end{abstract}


\section{Introduction}
Designing the layout of analog circuits is a crucial and complex task requiring significant expertise due to their susceptibility to noise, parasitics, alongside stringent topological requirements. This often leads to multiple iterations for layout engineers to achieve an optimal result.

The procedure involves two distinct yet closely entangled steps to place and connect circuit devices, called floorplanning and routing respectively.
Metaheuristics as simulated annealing (SA), particle swarm optimization (PSO), and genetic algorithms (GA) \cite{singh_review_2016} have been employed to streamline the floorplanning step. However, a primary drawback is that they cannot utilize past or external knowledge to enhance exploration of the solution space, as they optimize each problem instance anew.
Works as \cite{li_customized_2020} instead opted to leverage machine learning (ML) based solutions, specifically Graph Neural Networks (GNNs), to improve the generalization capabilities in generating optimal floorplans spanning across diverse circuit topologies. 
Reinforcement learning (RL) techniques have become effective in tackling combinatorial problems \cite{RL4CO}, such as floorplanning, by optimally navigating and focusing on the most promising solution space regions. Since floorplanning can be framed as a sequential decision-making process within a Markov Decision Process (MDP), RL techniques have led to state-of-the-art outcomes in several digital layout applications \cite{lai_maskplace_2022,cheng_joint_2021,amini_generalizable_2022,lai_chipformer_2023,yang_miracle_2024}. Nevertheless, the use of RL in \emph{analog} layout remains limited, calling for further exploration.

In this paper, we propose a combination of Relational Graph Convolutional Neural Networks (R-GCNs) \cite{schlichtkrull_modeling_2017} and RL to create optimal floorplans for diverse analog circuits types and topologies. The R-GCN model provides detailed circuit data information to the RL agent, which combines them with spatial encodings from a Convolutional Neural Network (CNN) \cite{oshea2015introduction} to determine the best shape and placement for each component. We also integrate this methodology with the ANAGEN procedural generator framework \cite{passerini_anagen_2021, demiri_anagen_2023} and an Obstacle Avoiding Rectilinear Steiner Tree (OARSMT) global router, as in \cite{basso_layout_2024}, streamlining the pipeline for automatic analog IC layout generation. The workflow overview is depicted in Figure \ref{fig:auto_layout_overview}.
\begin{figure}
    \centering
    \includegraphics[width=.97\columnwidth]{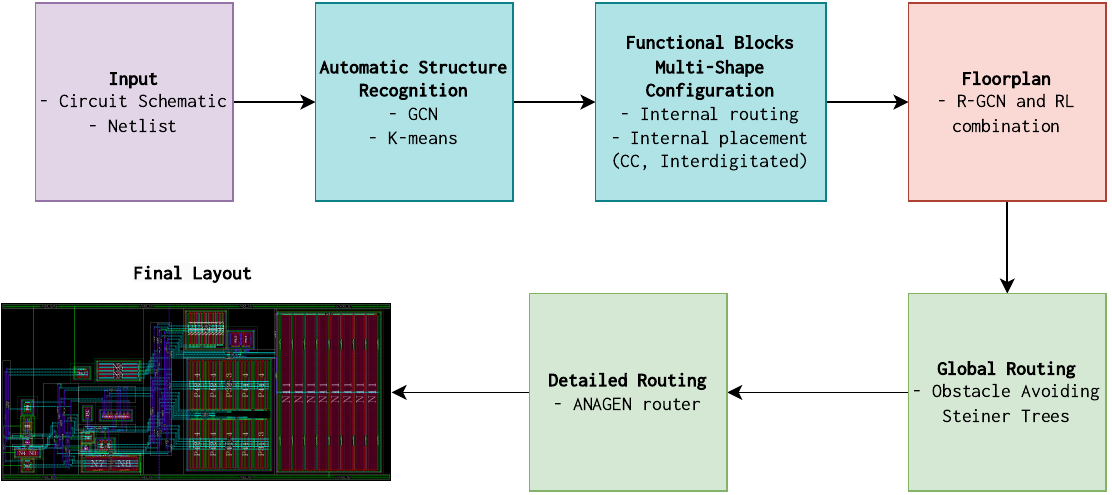}
    \caption{Overview of the automatic layout pipeline.}
    \label{fig:auto_layout_overview}
\end{figure}
The key contributions of this paper are as follows:
\begin{itemize}
    \item We train an R-GCN model to predict rewards for circuit placements. Once trained, it serves as an encoder of circuit, device, and geometric constraints for RL agent use to guarantee optimal generalization capabilities.
    \item We present an RL agent that combines graph and pixel-level representations to comprehensively describe circuit and problem characteristics. The agent policy is designed to learn to select \emph{optimal shapes and positions} for components while ensuring \emph{no overlaps and adherence to constraints} such as symmetry and alignment. To our knowledge, this is the first time such approach is proposed.
    \item We validate our novel method on circuits of increasing complexity. Our approach outperforms traditional metaheuristics and existing RL-based methods in terms of area, proxy wirelength, speed and cost metrics.
    \item Our enhanced procedural layout generation pipeline consistently shortens design timeframes while matching and even surpassing the quality of manually designed layouts.
\end{itemize}

The remainder of this paper is organized as follows. Section \ref{sec:Related_Works} presents previous works. Background of GNNs and RL are introduced in Section \ref{sec:Background}. Section \ref{sec:Methodology} details the problem setting and our R-GCN and RL-based floorplanning approach. Experimental results are provided in Section \ref{sec:Results}, while conclusions are drawn in Section \ref{sec:Conclusions}.

\section{Related Works}
\label{sec:Related_Works}
Floorplanning automation has extensively relied on metaheuristic techniques such as SA, GA or PSO that, combined with topological model like Sequence Pair (SP) \cite{balasa_symmetry_2000} or B*-Tree \cite{shunmugathammal_novel_2020} to describe component's relative placement, targeted to minimize an objective function through a stochastic search process. Although producing compact floorplans, these methods leave insufficient space for routing tracks, ultimately resulting in unusable layouts. Additionally, implementing geometric constraints is challenging and tends to increase the already lengthy optimization runtimes, especially with more complex circuits.

Recently, learning-based methodologies have emerged as promising alternatives. 
For instance, \cite{li_customized_2020} trained a GNN to predict analog ICs performance based on device placement, plugging it into an SA optimizer, but did not use it to directly produce a floorplan or consider positional constraints.
Gusm\~{a}o et al. in \cite{de_gusmao_scalable_2022} developed an unsupervised encoder-decoder model using attention mechanisms and R-GCNs to embed topological constraints (symmetry and proximity) and generate floorplans; yet, validation to remove overlaps is required and no routing related optimization metric is considered. 
Ahmadi et al. \cite{ahmadi_analog_2021} trained an RL agent to place FinFET modules on a grid, minimizing symmetry and alignment errors, area, and wirelength, but the assumption of fixed device shapes limited the approach's flexibility.
This work addresses the aforementioned limitations.

\section{Preliminaries}
\label{sec:Background}
\subsection{Graph Neural Networks}
Circuit netlists and layouts can be represented as graphs, leading EDA methods to benefit from applying GNNs to various design stages \cite{servadei_GNN_2021}. 
A graph is described as a tuple $G = (V,E)$, where $V$ is the set of nodes and $E$ is the set of edges. The neighborhood of node $v$, denoted as $\mathcal{N}(v)$, is $ u | (u,v) \in E$. A graph is \textit{directed} if edge direction matters; otherwise, it is \textit{undirected}. A graph with multiple types of nodes or edges is defined as \textit{heterogeneous}, whereas one with a single type is \textit{homogeneous}.
GNNs are deep learning models that natively operate on graph-structured inputs, aiming to learn continuous embedding vectors per node through a message passing process \cite{gilmer2017neural}. Nodes exchange vector-based messages with adjacent nodes over iterations, progressively enriching their state with neighborhood context. 
Graph Convolutional Neural Networks (GCNs) \cite{kipf_semi-supervised_2016} aggregate features as follows:
\begin{equation}
    h_u^{(l+1)} = \sigma \left(\sum_{v\in \mathcal{N}(u)}\frac{h_v^{(l)} W^{(l)}}{c_u}\right),
\end{equation}
being $h_u^{(l+1)}$ the updated node feature vector, $\sigma$ a non-linear differentiable function, $l$ the l-th GCN layer, $h_v^{(l)}$ the neighboring nodes feature vector, $W^{(l)} \in \mathbb{R}^{d \times d}$ a learnable weight matrix, and $c_u$ a normalization factor.
R-GCN extend the aggregation process to heterogeneous graphs, accounting for different node relationships:
\begin{equation}
    h_u^{(l+1)} = \sigma \left(W_0^{(l)}\cdot h_u^{(l)} + \sum_{r \in R} \sum_{v\in \mathcal{N}^r(u)} \frac{W_r^{(l)}\cdot h_v^{(l)}}{c_{u,r}}\right),
\end{equation}
where $\mathcal{N}^r(u)$ is the set of neighbors of node $u$ under relation $r \in R$, $W_0^{(l)} \in \mathbb{R}$ is a learnable weight vector for a node’s self-connection in each layer, and finally $c_{u,r}$ is again a normalization constant that varies with the R-GCN task.

\subsection{Reinforcement Learning}
RL techniques train an agent to discover effective solutions through interactions with the environment, utilizing an MDP framework denoted as $(\mathcal{S}, \mathcal{A}, \mathcal{P}, \mathcal{R}, \gamma)$.
At each time step $t$, the RL agent is in a state $s_t \in \mathcal{S}$ and chooses an action $a_t \in \mathcal{A}$ to execute. Following the transition probability $\mathcal{P}$, the agent transitions to a new state $s_{t+1}$ and receives a reward $r_t \in \mathcal{R}$ indicating the impact of its action, discounted by $\gamma$, to balance the relevance of immediate versus future rewards.
The agent eventually learns an optimal policy $\pi^*(a|s)$ that maximizes the expected sum of discounted rewards $G_t = \sum\limits_{k=0}^\infty \gamma^k R_{t+k+1}$.

\section{Methodology}
\label{sec:Methodology}
\subsection{Reinforcement Learning Formulation}
\label{subsec:problem_formulation}
Enhancing the floorplanning algorithm by Basso et al. in \cite{basso_layout_2024}, we propose an integration of an R-GCN model for chip representation learning with an RL agent. 
The agent selects a shape of a circuit components and places it on a discretized grid corresponding to the layout space. 
The floorplanning problem is formulated as the following MDP:
\begin{itemize}
    \item \textbf{States}: The state $s_t$ combines detailed information encoded by the R-GCN model, such as circuit graph $g$ and current instance $n_k$ $32$-dimensional embeddings, as well as local feature maps extracted at the pixel level by a CNN. The local feature maps consist of a $32{\times}32$ grid representation $f_g \in \{0,1\}^{32 \times 32}$, and two reward-related masks showing the increase in the placement empty space and wirelength proxy metric, similar to \cite{lai_maskplace_2022}, respectively denoted as $f_{ds}, f_w \in [0,1]^{32 \times 32}$. Additionally, there are three positional masks $f_{p} \in \{0, 1\}^{3\times32\times32}$, used also for action masking, delineating the admissible placement cells for the next block given the three possible shapes and adherence to non-overlapping and optionally defined spatial constraints.
    \item \textbf{Actions}: The action $a_t$ at time step $t$ consists in selecting one of three possible shapes for the current block $b_t$ and determining the grid cell to place its lower left corner.
    \item \textbf{Rewards}: We define a partial reward $r_t$ as the negative increase of proxy wirelength and empty space in the floorplan after $a_t$. The end of episode reward is instead equal to the negative weighted sum of floorplan's area, half-perimeter wirelength and, if specified, discrepancy w.r.t. the target aspect ratio. 
\end{itemize}

\subsection{Preliminary Structure Recognition (SR) and Functional Block Configuration}
\label{subsec:preliminary}
Given an input schematic, we use Infineon's GCN-based SR tool \cite{patel_machine_nodate} to detect circuit functional blocks.
Following \cite{basso_layout_2024}, we generate different block shapes by keeping a fixed total device width, i.e. area, and tailoring internal routing and device placement based on the recognized functional structure. These configurations are then provided to the RL agent.

\subsection{R-GCN Circuit Representation Learning}
An optimal floorplanning algorithm should be capable to generalize its performances across various circuit configurations and constraints. 
R-GCNs can effectively handle graphs of varying dimensions and topology, thanks to their permutation invariance.
Works as \cite{amini_generalizable_2022, mirhoseini_graph_2021, le_toward_2023} proposed to pre-train a GNN model on the supervised task of predicting rewards of input circuit graphs.
By aligning training tasks with the RL agent's goals, circuit embeddings produced for subsequent stages capture meaningful signals, enhancing RL agent's decision-making with augmented generalization capabilities.
In our setting, shown in Figure \ref{fig:circuit_graph}, circuits are represented as undirected graphs where each node $v_i$ corresponds to a functional block or single device. The edges $e_i$ represent relationships between nodes $(u,v)$, which can be connectivity (if they are connected in the netlist), horizontal or vertical alignment, or horizontal or vertical symmetry.
A node feature vector $x_u \in X$ includes the block area, internal parameters like transistor or resistance stripe width, terminals routing direction, pin counts, and a 28-dimensional one-hot encoding of the block's functional structure (e.g., current mirror, differential pair, cascode, etc.).

\begin{figure}
    \centering
    \includegraphics[width=.9\columnwidth]{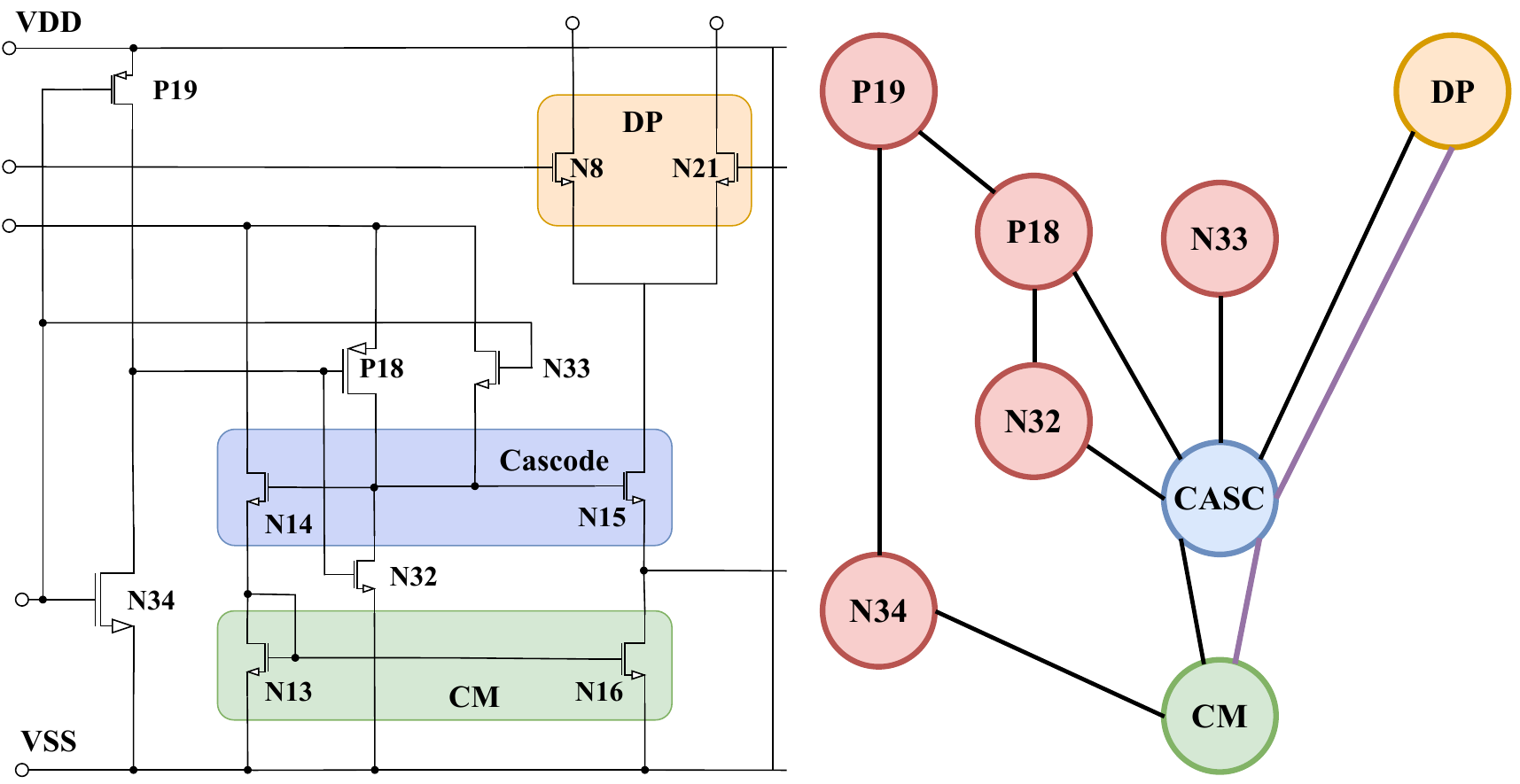}
    \caption{8-structure OTA circuit schematic with its graph representation. Violet edges are for vertical alignment and black for connectivity. Nodes are colored according to the functional block type.}
    \label{fig:circuit_graph}
\end{figure}

\subsubsection*{R-GCN Pre-Training Setup}
\begin{figure}
    \centering
    \includegraphics[width = \columnwidth]{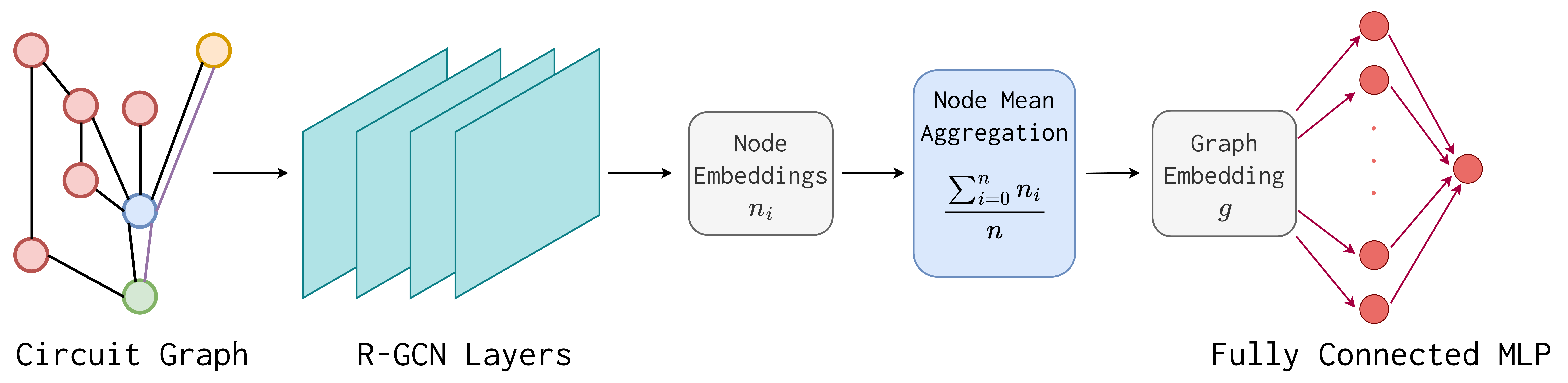}
    \caption{R-GCN architecture for circuit reward prediction.}
    \label{fig:rgcn_arch}
\end{figure}
Figure \ref{fig:rgcn_arch} shows the R-GCN model architecture, which consists of $4$ R-GCN layers followed by a node mean aggregation block to produce the whole graph embedding. This is then fed into $5$ fully connected (FC) layers to predict the reward value. 
The R-GCN training dataset comprises $21600$ floorplans and corresponding reward values, generated by optimizing placement w.r.t. area and proxy wirelength metrics using a mixture of SA, GA, and PSO. The circuits of interest vary in size and complexity, including operational transconductance amplifiers (OTAs), bias circuits, drivers, level shifters, clock synchronizers, comparators, and oscillators. Moreover, we ensured a balance between constrained and unconstrained floorplans. The supervised model is trained to minimize the mean squared error between the ground truth and predicted reward associated with the input circuit graph. 


\subsection{R-GCN and RL-based Floorplanning}
\begin{figure*}
    \centering
    \includegraphics[width = .935\textwidth]{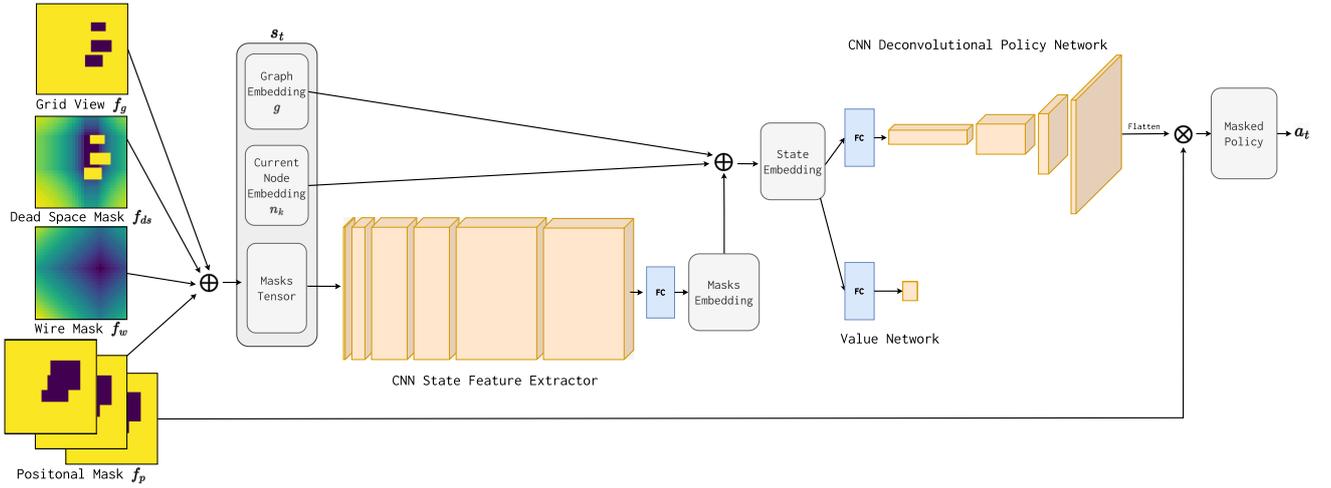}
    \caption{Overview of the RL model, enriched with CNN based feature extractor and policy network.}
    \label{fig:rgcn_RL_arch}
\end{figure*}

Given the R-GCN trained model, we remove the final FC layers and use the remaining part as encoder for the RL agent, enhancing the transfer learning capabilities of our methodology. Figure \ref{fig:rgcn_RL_arch} provides an overview of the RL architecture. We train the RL agent using a masked version of Proximal Policy Optimization (PPO) \cite{schulman_proximal_2017}, a state-of-the-art on-policy algorithm.
\subsubsection{Designing Action Space and Masking}
Large action spaces can hinder RL convergence towards an optimal policy. Therefore, we discretize the layout canvas into a $32{\times}32$ grid $f_g$, effectively balancing performance and accuracy by containing the action space while ensuring constraints satisfaction. The grid height $H$ and width $W$ are computed as $W{=}H{=}\sqrt{\frac{\sum_{i=1}^m A_i}{R_{\text{max}}}}$ being $A_i$ the area of the $i^{th}$ circuit device and $R_{\text{max}}{=}11$ the maximum empirically derived aspect ratio for a floorplan. Since both $H$ and $W$ depend on the size of each functional block, this design choice accommodates any complex circuit placement.
The agent can choose from $3$ candidate shapes for a circuit structure, similar to the flexibility human designers have. Combined with selecting the cell for placing the lower-left corner of a block, this results in an action space $\mathcal{A}$ of size $3{\times}32{\times}32{=}3072$.
To prevent further escalation of action space dimensionality, we use a heuristic inspired by \cite{mirhoseini_graph_2021}, which arranges block placement in order of decreasing size.\looseness=-1

As mentioned earlier, our floorplanning methodology can handle fundamental positional requirements such as symmetry, alignment, and guarantee the absence of device overlap. Works as \cite{huang_invalid_2022} prove that policy gradient algorithms can take advantage of masking procedures to avoid selecting invalid actions based on state information. Therefore, at each episode step, we generate three positional masks $f_{p}$, one for each possible candidate shape. 
These masks are obtained by combining two binary matrices: one representing partial placement and the other symmetry or alignment constraint masks. In the first one, a value of $1$ indicates an available cell, while $0$ signifies an already occupied one. The second one designates with $1\text{s}$ where the corresponding constraint would be satisfied (based on the placement of blocks belonging to a constraint group and the corresponding symmetry or alignment axis) and $0\text{s}$ where it would not. We map the real sizes of each circuit instance $(w, h)$ without approximation, where $w_g$ and $h_g$ are the respective scaled width and height on the grid: $$w_g=\ceil[\Bigg]{\frac{w\times32}{W}}, h_g= \ceil[\Bigg]{\frac{h\times32}{H}}.$$

\subsubsection{State Design}
Given that the R-GCN model does not supply detailed location information for specific circuit instances during an episode, we enrich the agent's state representation. As mentioned in Section \ref{subsec:problem_formulation}, this is achieved by combining the current node, i.e. block $b_t$, and graph, i.e. circuit, embeddings $n_k$ and $g$ with $6$ additional grid-based masks: $f_g$, $f_w$, $f_{ds}$ and $f_p$, which reflect partial placements, wire and dead space, i.e. empty space, increases, and valid positions. This augmented state aids the RL agent in identifying the optimal placement to optimize rewards.
Lai et al. \cite{lai_maskplace_2022} first proposed this approach, using $f_p$, $f_g$ and computing $f_w$ as the increase in proxy wirelength when placing $b_t$ in a specific position. Our approach extends the positional mask $f_p$ to account for multiple device shapes and introduces the dead space mask $f_{ds}$, a normalized, continuous matrix indicating the increase in empty space if $b_t$ is placed in a certain location. To construct $f_{ds}$, we iterate over all available cells on the placement grid, compute the resulting dead space from placing $b_t$, and subtract the previous dead space value. Already occupied locations are set to the maximum increment $1$ to mask invalid positions.
Figure \ref{fig:wire_dspace_masks} provides a visual example of $f_w$ and $f_{ds}$.
\begin{figure}
    \centering
    \includegraphics[width = .85\columnwidth]{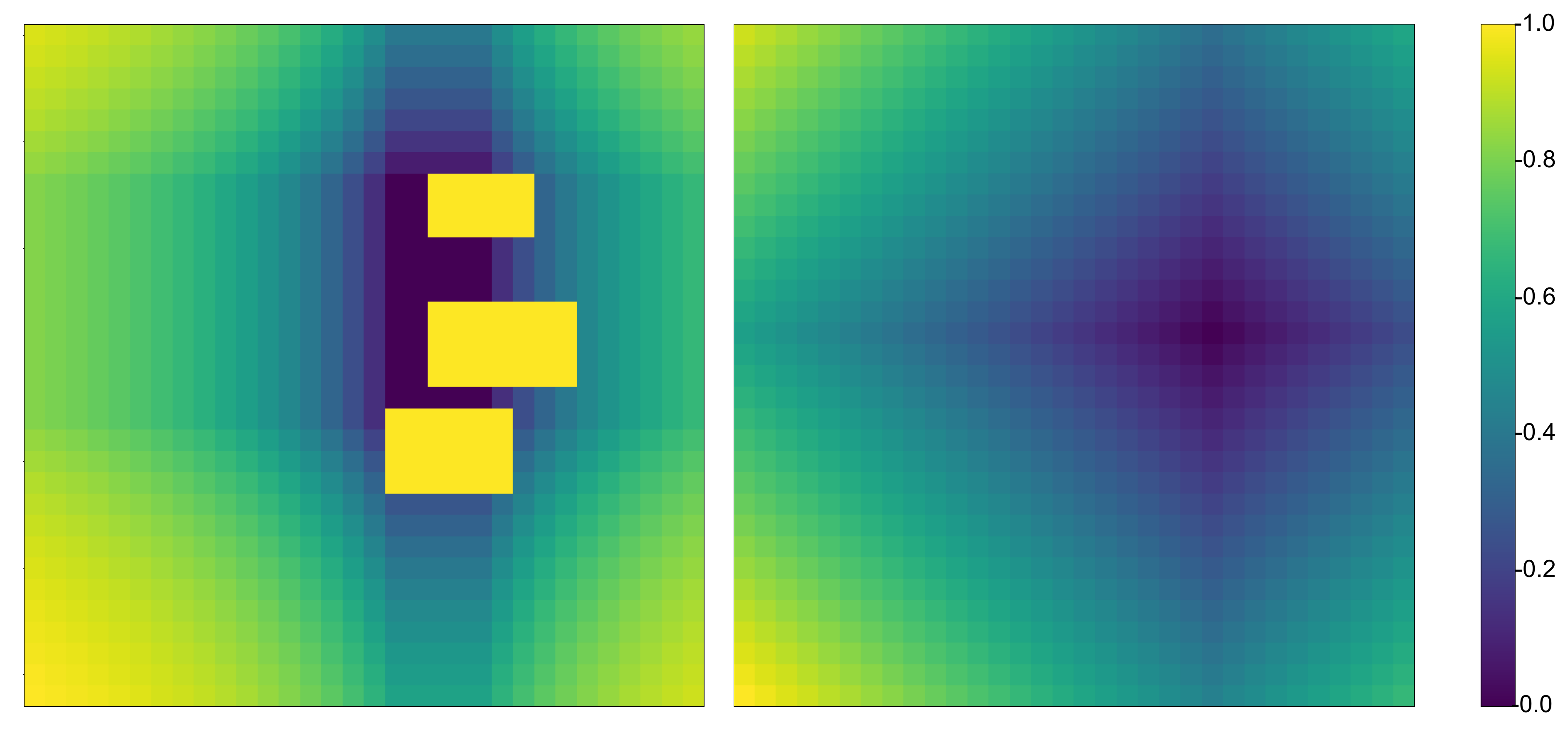}
    \caption{Dead space (left) and wire (right) masks. Darker areas highlight higher rewards regions in the case a block is placed there.}
    \label{fig:wire_dspace_masks}
\end{figure}

\subsubsection{Policy Design}
Treating grid masks as analogous to images, as suggested by \cite{cheng_joint_2021}, leverages CNNs' effectiveness in generating informative embeddings. 
For this reason, we concatenate the masks forming a single tensor of dimension $6{\times}32{\times}32$ and feed them to a CNN.
The convolution layers use a $3{\times}3$ kernel size with stride of $1$, padding of $1$ and $16,32,32,64,64$ filter channels, followed by one FC layer to produce a $512$-dimensional embedding vector.
Finally, after concatenating the R-GCN embedding and CNN outputs, this compact state representation is fed into both value and policy network. The policy includes a single FC layer that converts the input vector to a $512$-dimensional one, $3$ deconvolution layers with a kernel size of $4\times4$, a stride of $2$, a padding of $1$ and $32$, $16$, and $8$ filter channels. The policy then generates a probability distribution over actions, allowing the agent to jointly decide the shape and location of 
the new device to be placed.

\subsubsection{Reward Shaping}
As delineated in Section \ref{subsec:problem_formulation}, the agent's goal is to minimize two primary metrics: area occupation and half-perimeter wirelength (HPWL). HPWL, a widely used approximation for true wirelength, is computed as the half-perimeter of all nets' bounding boxes:
\begin{equation}
    \label{eq:HPWL}
    \text{HPWL} = \sum_{i=1}^n \text{max}(x_i)-\text{min}(x_i) + \text{max}(y_i)-\text{min}(y_i),
\end{equation}
where $x_i$, $y_i$ are the endpoints of a net, and $n$ is the total count of nets in the netlist. The dead space DS in a floorplan $F$ is computed as $1 - \frac{\sum_{i=1}^m A_i}{F_{area}}$.
To better guide the agent during an episode rollout, we provide intermediate rewards $r_t$ based on the increase in partial floorplan dead space and HPWL, computed from the currently placed instances after possibly taking action $a_t$. The intermediate reward is defined as: 
\begin{equation}
    \label{eq:partial_reward}
    r_t = -(\Delta_{ds} + \Delta_{\text{HPWL}}),
\end{equation}
where $\Delta_{ds}=\text{DS}_t - \text{DS}_{t-1}$ and $\Delta_{\text{HPWL}}=\text{HPWL}_t - \text{HPWL}_{t-1}$.
Given the optional constraint of a fixed aspect ratio for the final floorplan, we define the agent's end of episode reward $\mathcal{R}$ as:
\begin{equation}
    \label{eq:reward_func}
    \mathcal{R} = -\left( \alpha \frac{F_\text{area}}{\sum_{i=1}^m A_i} + \beta \frac{\text{HPWL}}{\text{HPWL\textsubscript{min}}}  + \gamma(R^* - R)^2 \right).
\end{equation}
Here, $\alpha$, $\beta$ and $\gamma$ are empirically set weights ($1$, $5$, and $5$, respectively) determined through extensive experimentation. These weights balance area, wirelength and fixed outline error terms in terms of their magnitude and impact on the final floorplan quality. $\text{HPWL\textsubscript{min}}$ is the minimum HPWL value estimated through a metaheuristic-based simulation for standardization while $R^*$ and $R$ are the target and current floorplan aspect ratios, respectively. Finally, whenever the generated floorplan violates any predefined constraint, we penalize the agent's behavior with a reward of $-50$.

\subsubsection{RL Training Schedule}
\label{subsec:RL_training}
Our methodology is designed to develop a single RL agent capable of generating optimal floorplans for a wide array of circuit types and constraints. To achieve this, we use a hybrid curriculum learning (HCL) approach, described in \cite{zaremba_learning_2015}, which incrementally presents the agent circuits of growing complexity. Starting with smaller circuits, we interleave them with random sampling of new circuit instances and constraints. This method maintains the agent's exposure to complex scenarios and prevents the loss of previously acquired knowledge, thereby enhancing transferability to new, unseen instances. 
The circuits used to train the RL are $3$ operational transconductance amplifiers (OTAs) and $2$ bias ones, respectively encompassing $3,5,8,3\text{ and }9$ blocks, to ensure enough diversity in the data.

\subsection{Routing and Final Layout Generation}
Once a feasible floorplan is generated, we construct an OARSMT for each net to minimize wirelength and avoid obstacles. Unlike \cite{basso_layout_2024}, which required congestion estimation to reserve space for routing channels, our method yields routing ready floorplans. The global routing tree is segmented into conduits, detailing connections and layers, guiding ANAGEN's router to finalize circuit connections.
This approach enhances ANAGEN's capabilities, potentially making it competitive with current state-of-the-art techniques \cite{magical, dhar_align_2020}, while preserving the correctness inherent in procedural generation and leveraging RL for advanced floorplan generation.

\section{Experimental Results}
\label{sec:Results}
Our floorplanning pipeline is built in Python 3.9, using DGL \cite{wang2019dgl} and Stable Baselines3 \cite{stable-baselines3} libraries for implementing respectively the R-GCN and RL models. 

\subsection{RL Placement Training Setup}
\begin{figure}
    \centering
    \includegraphics[width = .93\columnwidth]{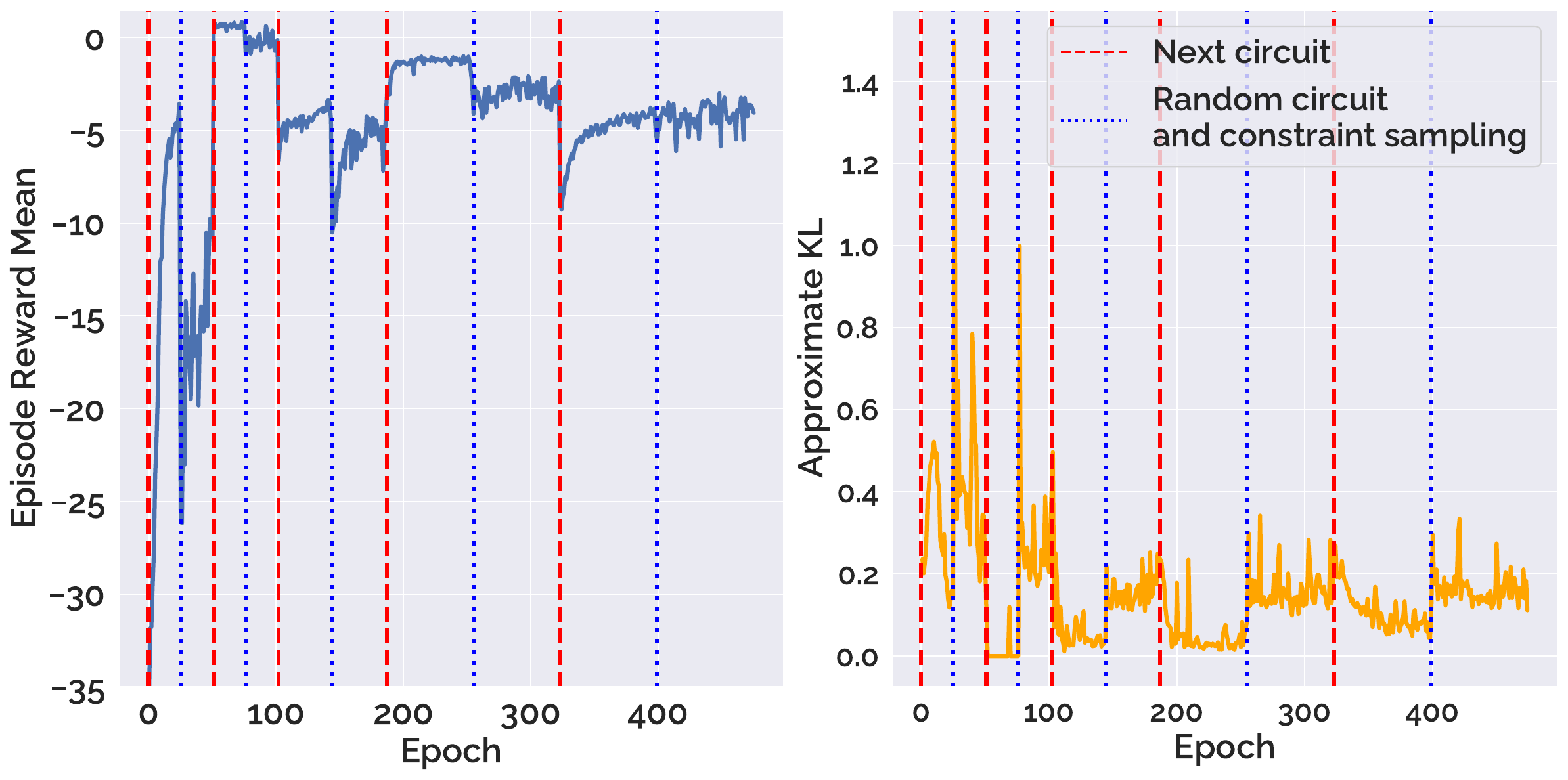}
    \caption{Episode Reward Mean and Approximate Kullback-Leibler divergence during RL agent's HCL training schedule.}
    \label{fig:rollout_mean}
\end{figure}
To foster policy robustness and reduce convergence time, we use $16$ parallel environments to gather multiple experiences.
Following the HCL approach from Section \ref{subsec:RL_training}, we train each circuit for $4096$ episodes. During the first half of the episodes we avoid introducing new constraints or circuits. After this phase, we begin sampling new circuit instances and constraints with probabilities $p_{\text{circuit}}=0.5 \text{ and } p_{\text{constraint}}=0.3$, respectively.
Figure \ref{fig:rollout_mean} illustrates the benefits of the HCL procedure for convergence towards a stable policy. The agent can recover optimal rewards and achieve low approximate Kullback-Leibler divergence values during circuit sampling, indicating strong generalization and robustness. Training the RL agent on a single Nvidia A30 GPU took 12 hours and 42 minutes.

\subsection{Comparison Against Baselines}
\begin{table*}
\centering
\caption{Comparative Analysis of R-GCN and RL Method, Across Various Fine-Tuned Setups, Versus Previous Techniques.}
\label{tab:GNNRLvsOthers}
\resizebox{\textwidth}{!}{%
\begin{tabular}{@{}cccccccccccc@{}}
\toprule
Circuit & \# Struct. & Metric & R-GCN RL 0-shot & R-GCN RL 1-shot & R-GCN RL 100-shot & R-GCN RL 1000-shot & SA & GA & PSO & RL-SA \cite{basso_layout_2024} & RL \cite{basso_layout_2024} \\ \midrule
 &  & Runtime (s) & \textbf{0.06 $\pm$ 0.24} & 7.97 $\pm$ 0.31 & 15.88 $\pm$ 0.3 & 174.13 $\pm$ 3.55 & 0.91 $\pm$ 0.01 & 4.58 $\pm$ 0.01 & 6.86 $\pm$ 0.03 & 1.03 $\pm$ 0.01 & 53.08 $\pm$ 0.66 \\
 &  & Dead space (\%) & 57.5 $\pm$ 1.25 & 47.19 $\pm$ 1.7 & 47.38 $\pm$ 4.32 & \textbf{43.93 $\pm$ 1.82} & 49.79 $\pm$ 4.28 & 53.95 $\pm$ 5.74 & 46.75 $\pm$ 3.16 & 51.19 $\pm$ 4.2 & 49.05 $\pm$ 3.93 \\
 &  & HPWL (\textmu m) & 150.62 $\pm$ 9.0 & 196.89 $\pm$ 11.03 & 175.94 $\pm$ 18.61 & \textbf{131.34 $\pm$ 23.76} & 166.03 $\pm$ 18.72 & 175.05 $\pm$ 25.85 & 164.32 $\pm$ 11.05 & 164.17 $\pm$ 17.81 & 166.96 $\pm$ 19.75 \\
\multirow{-4}{*}{OTA-1} & \multirow{-4}{*}{5} & Reward & -2.37 $\pm$ 0.42 & -3.35 $\pm$ 0.46 & -2.61 $\pm$ 0.87 & \textbf{-0.21 $\pm$ 0.91} & -2.04 $\pm$ 0.68 & -2.46 $\pm$ 1.13 & \textit{\underline{-1.86 $\pm$ 0.47}} & \underline{-1.97 $\pm$ 0.67} & -2.05 $\pm$ 0.82 \\ \midrule
 &  & Runtime (s) & \textbf{0.14 $\pm$ 0.01} & 9.2 $\pm$ 0.09 & 28.0 $\pm$ 0.2 & 287.03 $\pm$ 0.88 & 1.09 $\pm$ 0.03 & 4.98 $\pm$ 0.02 & 7.30 $\pm$ 0.03 & 1.3 $\pm$ 0.1 & 94.64 $\pm$ 1.8 \\
 &  & Dead space (\%) & 43.77 $\pm$ 5.27 & \textbf{33.18 $\pm$ 1.95} & \textbf{33.18 $\pm$ 2.34} & 35.57 $\pm$ 1.62 & 57.5 $\pm$ 4.08 & 57.3 $\pm$ 6.77 & 55.3 $\pm$ 4.66 & 55.68 $\pm$ 4.48 & 54.17 $\pm$ 5.4 \\
 &  & HPWL (\textmu m) & 202.19 $\pm$ 22.89 & \textbf{154.27 $\pm$ 5.79} & 164.14 $\pm$ 9.37 & 168.29 $\pm$ 7.52 & 244.42 $\pm$ 41.68 & 237.3 $\pm$ 30.17 & 226.27 $\pm$ 31.32 & 229.17 $\pm$ 23.26 & 234.95 $\pm$ 37.39 \\
\multirow{-4}{*}{OTA-2} & \multirow{-4}{*}{8} & Reward & -2.52 $\pm$ 0.89 & \textbf{-0.68 $\pm$ 0.23} & \textit{\underline{-0.96 $\pm$ 0.31}} & \underline{-1.16 $\pm$ 0.28} & -3.97 $\pm$ 1.54 & -3.68 $\pm$ 1.33 & -3.22 $\pm$ 1.17 & -3.51 $\pm$ 0.85 & -3.62 $\pm$ 1.38 \\ \midrule
 &  & Runtime (s) & \textbf{0.16 $\pm$ 0.15} & 8.96 $\pm$ 0.51 & 26.17 $\pm$ 0.22 & 303.62 $\pm$ 3.13 & 1.23 $\pm$ 0.01 & 5.4 $\pm$ 0.37 & 7.74 $\pm$ 0.48 & 1.57 $\pm$ 0.12 & 72.06 $\pm$ 1.27 \\
 &  & Dead space (\%) & 53.16 $\pm$ 3.97 & 56.85 $\pm$ 1.54 & 59.93 $\pm$ 5.64 & \textbf{45.52 $\pm$ 7.22} & 67.79 $\pm$ 4.74 & 73.28 $\pm$ 5.06 & 68.22 $\pm$ 4.77 & 67.57 $\pm$ 4.29 & 67.58 $\pm$ 4.68 \\
 &  & HPWL (\textmu m) & 271.51 $\pm$ 39.03 & 313.63 $\pm$ 23.50 & 321.76 $\pm$ 24.11 & \textbf{191.91 $\pm$ 64.07} & 288.97 $\pm$ 40.05 & 269.7 $\pm$ 47.05 & 329.14 $\pm$ 55.57 & 285.19 $\pm$ 39.65 & 288.46 $\pm$ 39.74 \\
\multirow{-4}{*}{Bias-1} & \multirow{-4}{*}{9} & Reward & \textit{\underline{-5.68 $\pm$ 1.51}} & -7.40 $\pm$ 0.87 & -7.70 $\pm$ 0.83 & \textbf{-2.53 $\pm$ 2.60} & -6.61 $\pm$ 1.7 & \underline{-6.29 $\pm$ 2.25} & -8.07 $\pm$ 2.29 & -6.5 $\pm$ 1.69 & -6.65 $\pm$ 1.65 \\ \midrule
\rowcolor[HTML]{EFEFEF} 
\cellcolor[HTML]{EFEFEF} & \cellcolor[HTML]{EFEFEF} & Runtime (s) & \textbf{0.11 $\pm$ 0.0} & 6.82 $\pm$ 1.09 & 17.65 $\pm$ 0.05 & 166.22 $\pm$ 0.75 & 0.99 $\pm$ 0.01 & 4.8 $\pm$ 0.01 & 6.71 $\pm$ 0.03 & 1.16 $\pm$ 0.03 & 42.96 $\pm$ 0.6 \\
\rowcolor[HTML]{EFEFEF} 
\cellcolor[HTML]{EFEFEF} & \cellcolor[HTML]{EFEFEF} & Dead space (\%) & 57.85 $\pm$ 2.54 & 53.17 $\pm$ 7.73 & 59.87 $\pm$ 2.25 & \textbf{33.76 $\pm$ 4.41} & 65.62 $\pm$ 4.32 & 69.99 $\pm$ 5.56 & 65.57 $\pm$ 4.27 & 63.84 $\pm$ 4.57 & 61.56 $\pm$ 3.63 \\
\rowcolor[HTML]{EFEFEF} 
\cellcolor[HTML]{EFEFEF} & \cellcolor[HTML]{EFEFEF} & HPWL (\textmu m) & 108.0 $\pm$ 3.22 & 118.06 $\pm$ 13.34 & 128.63 $\pm$ 16.53 & \textbf{97.55 $\pm$ 7.58} & 127.99 $\pm$ 17.01 & 128.26 $\pm$ 15.95 & 127.92 $\pm$ 16.62 & 124.11 $\pm$ 15.17 & 120.22 $\pm$ 15.1 \\
\rowcolor[HTML]{EFEFEF} 
\multirow{-4}{*}{\cellcolor[HTML]{EFEFEF}RS Latch} & \multirow{-4}{*}{\cellcolor[HTML]{EFEFEF}7} & Reward & \textit{\underline{ -4.04 $\pm$ 0.31}} & \underline{-4.63 $\pm$ 1.27} & -5.55 $\pm$ 1.26 & \textbf{-2.34 $\pm$ 0.62} & -5.44 $\pm$ 1.58 & -5.58 $\pm$ 1.59 & -5.39 $\pm$ 1.56 & -5.03 $\pm$ 1.44 & -4.69 $\pm$ 1.39 \\ \midrule
\rowcolor[HTML]{EFEFEF} 
\cellcolor[HTML]{EFEFEF} & \cellcolor[HTML]{EFEFEF} & Runtime (s) & \textbf{0.25 $\pm$ 0.0} & 11.6 $\pm$ 0.1 & 79.64 $\pm$ 0.81 & 814.81 $\pm$ 7.76 & 1.85 $\pm$ 0.01 & 6.63 $\pm$ 0.06 & 10.4 $\pm$ 0.06 & 2.24 $\pm$ 0.07 & 155.08 $\pm$ 3.53 \\
\rowcolor[HTML]{EFEFEF} 
\cellcolor[HTML]{EFEFEF} & \cellcolor[HTML]{EFEFEF} & Dead space (\%) & 63.5 $\pm$ 4.79 & 62.49 $\pm$ 1.73 & 59.71 $\pm$ 3.35 & \textbf{48.17 $\pm$ 4.11} & 71.44 $\pm$ 7.22 & 73.3 $\pm$ 7.16 & 70.64 $\pm$ 6.24 & 70.8 $\pm$ 7.82 & 69.36 $\pm$ 8.42 \\
\rowcolor[HTML]{EFEFEF} 
\cellcolor[HTML]{EFEFEF} & \cellcolor[HTML]{EFEFEF} & HPWL (\textmu m) & 1794.5 $\pm$ 173.3 & 1811.61 $\pm$ 163.9 & 1811.04 $\pm$ 133.7 & \textbf{1419.57 $\pm$ 61.88} & 1981.07 $\pm$ 319.51 & 2192.15 $\pm$ 448.45 & 2152.62 $\pm$ 294.24 & 1862.99 $\pm$ 372.89 & 1941.85 $\pm$ 516.42 \\
\rowcolor[HTML]{EFEFEF} 
\multirow{-4}{*}{\cellcolor[HTML]{EFEFEF}Driver} & \multirow{-4}{*}{\cellcolor[HTML]{EFEFEF}17} & Reward & \underline{-7.43 $\pm$ 1.29} & \underline{-7.43 $\pm$ 1.05} & \textit{\underline{-7.26 $\pm$ 0.84}} & \textbf{-4.43 $\pm$ 0.53} & -8.55 $\pm$ 2.41 & -10.23 $\pm$ 3.23 & -9.66 $\pm$ 2.23 & -7.69 $\pm$ 2.72 & -8.31 $\pm$ 3.73 \\ \midrule
\rowcolor[HTML]{EFEFEF} 
\cellcolor[HTML]{EFEFEF} & \cellcolor[HTML]{EFEFEF} & Runtime (s) & \textbf{0.34 $\pm$ 0.04} & 12.74 $\pm$ 1.4 & 88.73 $\pm$ 1.01 & 849.86 $\pm$ 3.56 & 2.07 $\pm$ 0.01 & 6.91 $\pm$ 0.02 & 11.6 $\pm$ 0.14 & 2.42 $\pm$ 0.02 & 244.62 $\pm$ 3.27 \\
\rowcolor[HTML]{EFEFEF} 
\cellcolor[HTML]{EFEFEF} & \cellcolor[HTML]{EFEFEF} & Dead space (\%) & 68.49 $\pm$ 6.81 & 56.91 $\pm$ 4.39 & 57.36 $\pm$ 3.22 & \textbf{45.12 $\pm$ 2.66} & 73.68 $\pm$ 4.2 & 70.36 $\pm$ 4.71 & 69.32 $\pm$ 5.75 & 74.89 $\pm$ 4.76 & 70.88 $\pm$ 6.37 \\
\rowcolor[HTML]{EFEFEF} 
\cellcolor[HTML]{EFEFEF} & \cellcolor[HTML]{EFEFEF} & HPWL (\textmu m) & 3375.84 $\pm$ 235.86 & 2967.92 $\pm$ 174.56 & 2780.59 $\pm$ 225.7 & \textbf{2141.84 $\pm$ 150.64} & 2896.52 $\pm$ 177.68 & 3501.24 $\pm$ 490.11 & 3346.66 $\pm$ 498.34 & 2872.94 $\pm$ 369.75 & 2854.06 $\pm$ 366.79 \\
\rowcolor[HTML]{EFEFEF} 
\multirow{-4}{*}{\cellcolor[HTML]{EFEFEF}Bias-2} & \multirow{-4}{*}{\cellcolor[HTML]{EFEFEF}19} & Reward & -5.95 $\pm$ 0.67 & \underline{-4.34 $\pm$ 0.7} & \textit{\underline{-3.65 $\pm$ 0.57}} & \textbf{-1.43 $\pm$ 0.47} & -5.17 $\pm$ 0.68 & -6.26 $\pm$ 1.65 & -5.74 $\pm$ 1.67 & -5.08 $\pm$ 1.3 & -4.36 $\pm$ 1.34 \\ \bottomrule
\end{tabular}%
}
\end{table*}
We evaluate our approach against established metaheuristics such as SA, the same methodology used by state-of-the-art automatic layout generator \cite{dhar_align_2020}, PSO, GA as well as the two methods from \cite{basso_layout_2024}, involving a combination of RL with SA and a pure RL technique both based on SP. Unfortunately, the circuits of interest are incompatible with other fully automated layout generation engines \cite{magical, dhar_align_2020} due to technology constraints.
Performances are measured on $3$ industrial designs from the RL training set and $3$ unseen ones, aiming to validate transferability with zero-shot and few-shot fine-tuning on novel circuits.
Few-shot learning involves refining a pre-trained RL agent by continuing its training on a specific problem or circuit instance, optimizing it for that context.
For a fair comparison, congestion-aware device spacing is applied to all other approaches to allocate sufficient room for routing channels, as our methodology provides routing-ready floorplans. No constraints are imposed on any circuit.

Table \ref{tab:GNNRLvsOthers} lists the interquartile mean and standard deviation values for algorithm computational runtime, floorplan HPWL, dead space, and associated reward. We emphasize the best results in bold, and highlight the second and third best rewards in italic underlined and underlined only, respectively. Unseen circuits are represented as grey rows.
As shown, our novel approach surpasses the baselines in terms of floorplan reward in all scenarios with proper fine-tuning, and in $4$ out of $6$ cases at zero-shot while significantly improving runtime. Moreover, our methodology demonstrates remarkable transfer capability across unseen and more complex designs, even without additional fine-tuning. On new circuits, HPWL and dead space percentage are reduced by $38.7\%$ and $66.8\%$, respectively, compared to past techniques. Moreover, few-shot fine-tuning improves results compared to the zero-shot model for the same number of iterations. 
While training time is significant for few-shot solutions, once fine-tuned, the model doesn't require retraining, making single floorplan generation runtimes comparable to zero-shot ones and outperforming other approaches.\looseness=-1

\subsection{Evaluation of Complete Layouts}
\begin{figure}
\centering
\subfloat[]{\includegraphics[width=0.34\columnwidth]{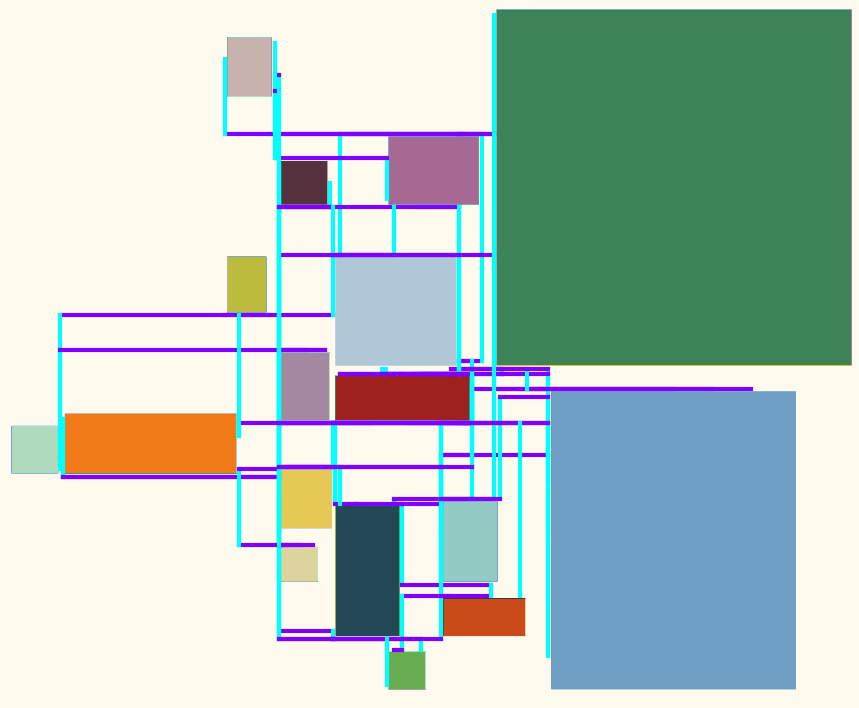}\label{fig:abstract_out}}\hfill
\subfloat[]{\includegraphics[width=0.32\columnwidth]{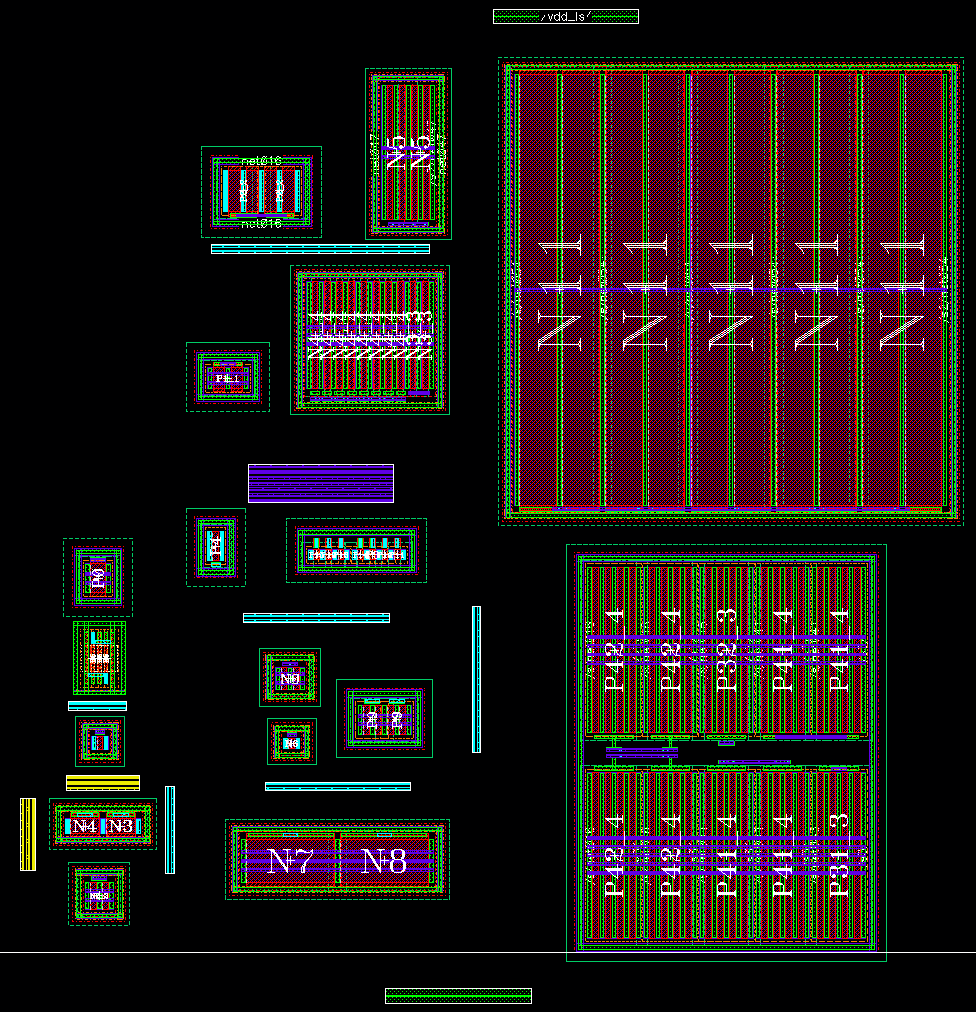}\label{fig:template_RL}}\hfill
\subfloat[]{\includegraphics[width=0.33\columnwidth]{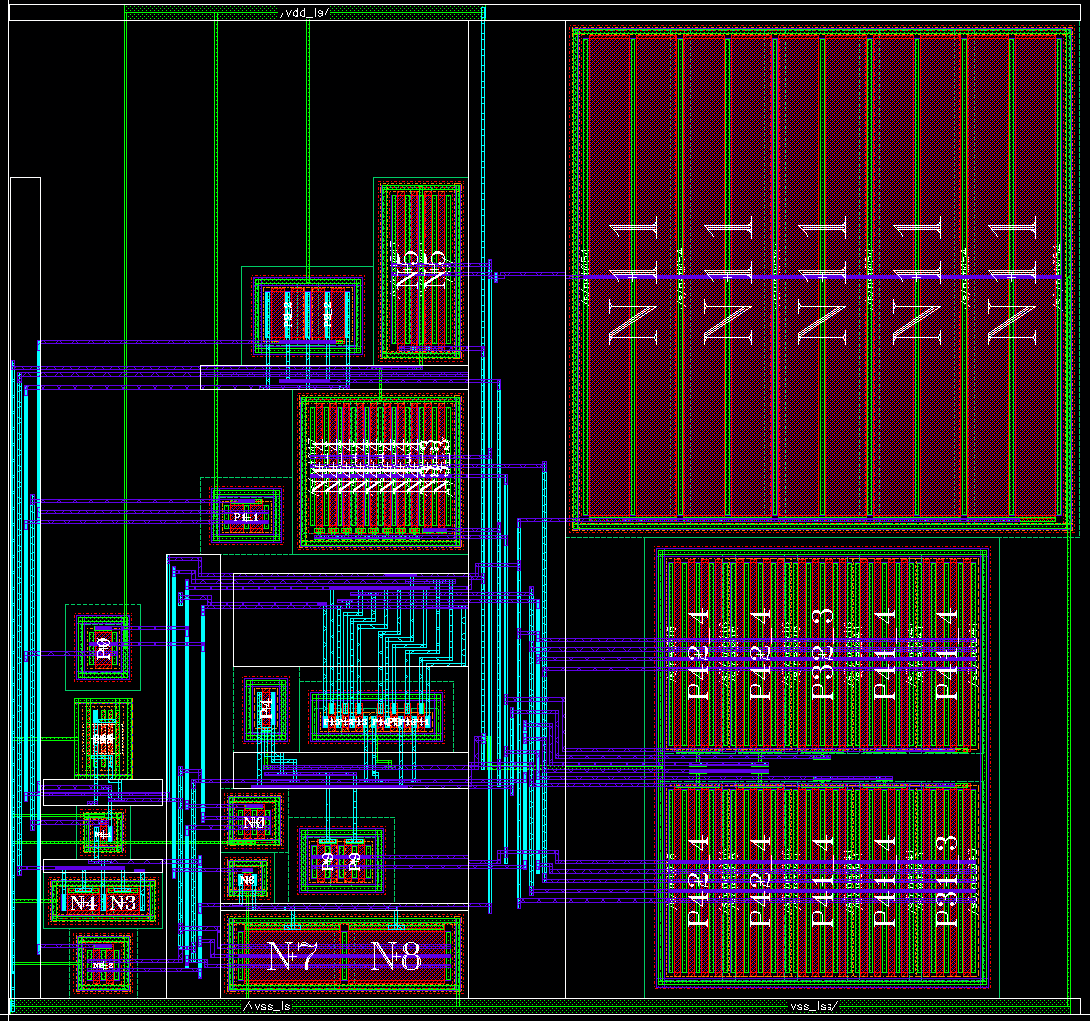}\label{fig:layout_RL}}
\vspace{-7pt}
\subfloat[]{\includegraphics[width=0.49\columnwidth]{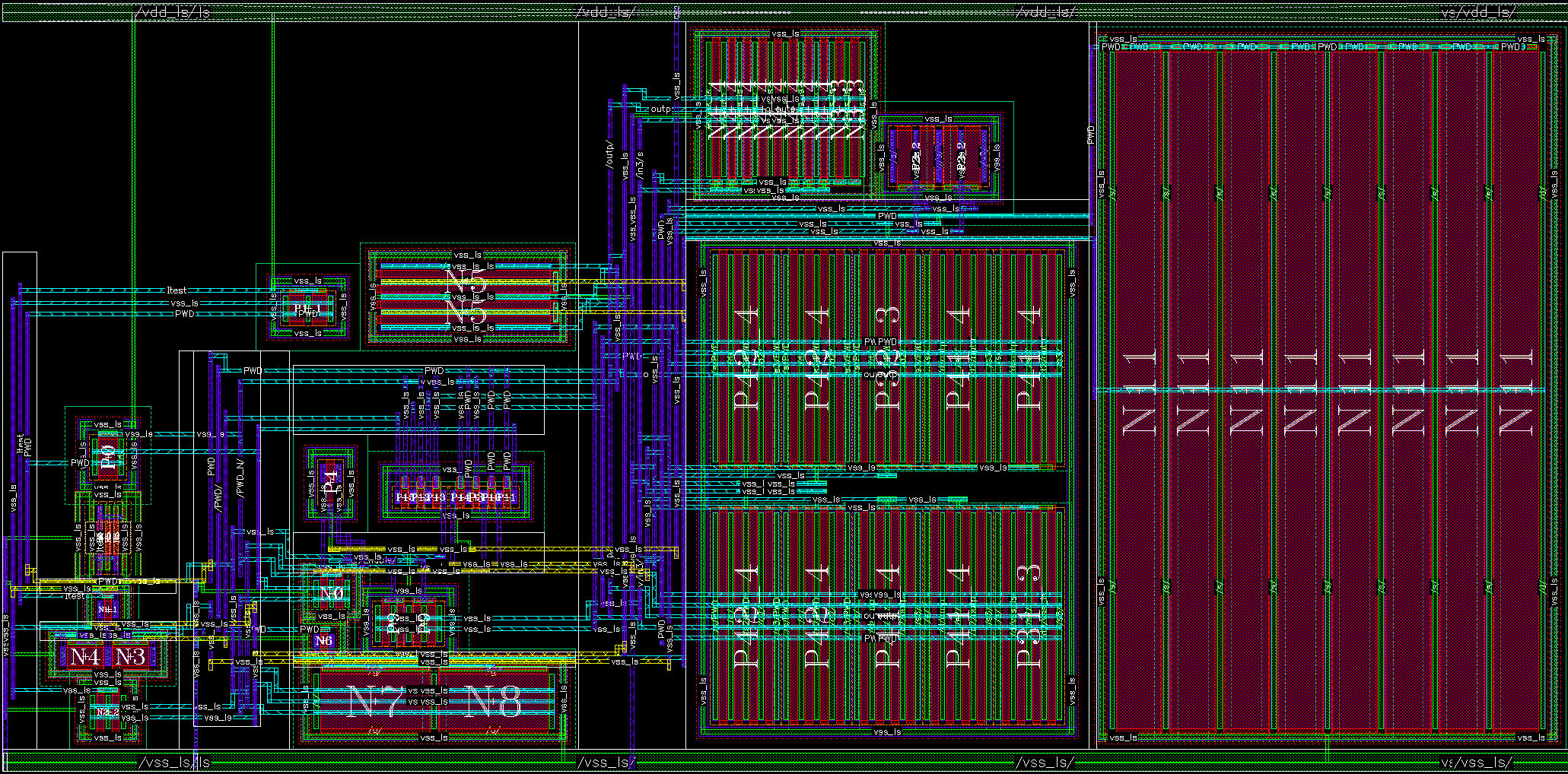}\label{fig:RL_opt_driver}}%
\hspace*{0.006\columnwidth}%
\subfloat[]{\includegraphics[width=0.5\columnwidth]{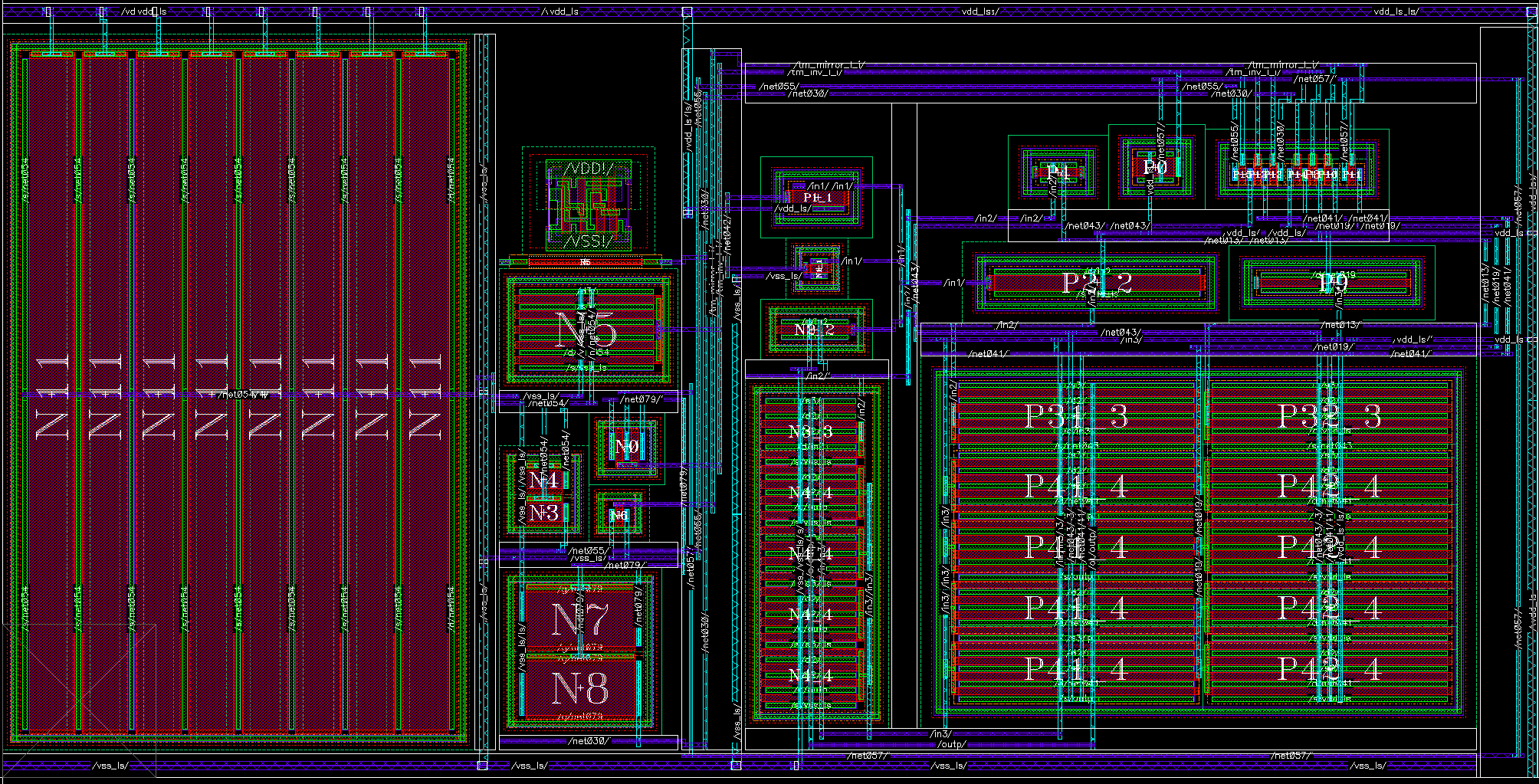}\label{fig:manual_driver}}
\caption{(a) RL-generated placement and OARSMT global routing, (b) post-adjustment floorplan and channels definitions, (c) resulting layout, (d) post-manual refinement optimized layout, and (e) full manual design.}\label{fig:Driver}
\end{figure}

\begin{table}
\centering
\caption{Comparison of Area, Dead Space, and Layout Generation Time Between Our Automated Method and Manual Design.}
\label{tab:practical_examples}
\resizebox{\columnwidth}{!}{%
\begin{tabular}{@{}ccccccc@{}}
\toprule
Circuit &
  Method &
  Area ($\text{\textmu m}^2$) &
  Dead space (\%) &
  \begin{tabular}[c]{@{}c@{}}Template \\ generation \\ time (s)\end{tabular} &
  \begin{tabular}[c]{@{}c@{}}Manual \\ improvement \\ time (h)\end{tabular} &
  \begin{tabular}[c]{@{}c@{}}Final layout \\ generation \\ time (h)\end{tabular} \\ \midrule
\multirow{2}{*}{OTA}  & Ours   & \textbf{228.6 (-14.1\%)}  & \textbf{30.01 (-5.98\%)} & 111.0 & 0.17 & \textbf{0.20 (-97.5\%)}  \\
                        & Manual & 266.0                     & 31.92                    & -     & -    & 8                        \\ \midrule
\multirow{2}{*}{Bias-1} & Ours   & 515.6 (+52.1\%)           & 54.01 (+8.68\%)          & 127.1 & 1    & \textbf{1.04 (-87.0\%)}  \\
                        & Manual & \textbf{247.1}            & \textbf{49.32}           & -     & -    & 8                        \\ \midrule
\multirow{2}{*}{Driver} & Ours   & \textbf{3584.7 (-2.43\%)} & \textbf{38.78 (-3.82\%)} & 456.3 & 20   & \textbf{20.13 (-37.1\%)} \\
                        & Manual & 3674.0                    & 40.32                    & -     & -    & 32                       \\ \bottomrule
\end{tabular}%
}
\end{table}
We plug our novel floorplanning algorithm into the pipeline proposed in \cite{basso_layout_2024} and validate its effectiveness by comparing the completed layouts of a 3-block OTA, 9-block Bias, and 17-block Driver from \cite{demiri_anagen_2023} against their manually designed counterparts. Notably, the manual Bias layout was crafted without using ANAGEN, unlike the other two.
The metrics of interest involve floorplan's area, dead space, and time required to produce a DRC and LVS clean layout. Results in Table \ref{tab:practical_examples} underscore algorithm's capability to quickly generate valid floorplans, facilitating faster design iteration.
The OTA and Driver circuits exhibit improved area and dead space metrics, although the Bias circuit has higher dead space percentages. This latter layout in fact benefits from the absence of ANAGEN's routing constraints, resulting in smaller area occupation.
Finally, Figure \ref{fig:Driver} shows a direct comparison between the automatically generated Driver layout from our methodology and the manually designed one. Figures \ref{fig:abstract_out} and \ref{fig:template_RL} illustrate the outcome of our floorplanning and global routing algorithm, providing clear guidance for physical design engineers on the expected wireflow.
In more complex layouts, manual refinement of routing channels guided by the OARSMT is still necessary to accommodate ANAGEN's routing sensitivities. However, for simpler layouts like the OTA example, routing channel generation was fully automated. Improving this aspect remains a focus for future work.\looseness=-1

\section{Conclusions and Future Research}
\label{sec:Conclusions}
This paper proposes a combined R-GCN and RL-based methodology for analog ICs floorplanning, ensuring alignment, symmetry, and no overlap constraints compliance. Our approach can generalize and transfer knowledge across different type of circuits, including new, unseen ones. The generated floorplans not only outperform established baselines but also, when integrated into a procedural generation pipeline, yield complete layouts of comparable quality to human generated ones in significantly reduced runtime. In the future, we aim to augment the floorplan algorithm with detailed routing information to further condition device placement towards easier and more efficient routing configurations.

\clearpage

\bibliographystyle{IEEEtran} 
\bibliography{IEEEabrv,paper_body/biblio}

\end{document}